\documentclass[10pt,twocolumn,letterpaper]{article}
\usepackage{stfloats}
\usepackage{algorithm}
\usepackage{listings}

\usepackage{amsthm,amsmath,amssymb}
\usepackage{pifont}

\usepackage{mathrsfs}
\usepackage[utf8]{inputenc} 
\usepackage[T1]{fontenc}    
\usepackage{booktabs}       
\usepackage{amsfonts}       
\usepackage{nicefrac}       
\usepackage{microtype}      
\usepackage{lipsum}
\usepackage{graphicx}
\graphicspath{ {./images/} }

\usepackage{subcaption}

\usepackage{makecell}

\usepackage[utf8]{inputenc} 
\usepackage[T1]{fontenc}    
\usepackage{booktabs}       
\usepackage{amsfonts}       
\usepackage{nicefrac}       
\usepackage{microtype}      
\usepackage{tabularx}
\usepackage{subcaption}
\usepackage{multicol}
\usepackage{multirow}
\usepackage{tabularray}
\usepackage{amsmath}
\usepackage{amssymb}
\usepackage{siunitx}
\usepackage{colortbl}
\usepackage{graphicx}
\usepackage{lipsum}

\usepackage[export]{adjustbox}
\usepackage{wrapfig}
\usepackage{svg}
\usepackage{amssymb}
\usepackage{sidecap}
\usepackage{caption}
\usepackage{tabularx} 
\usepackage{ragged2e} 
\usepackage{booktabs} 
\newcommand{\xmark}{\textcolor{red}{\ding{55}}}
\newcommand{\cmark}{\textcolor{green}{\ding{51}}}
\usepackage{wrapfig}

\usepackage[pagenumbers]{cvpr} 

\definecolor{cvprblue}{rgb}{0.21,0.49,0.74}
\usepackage[pagebackref,breaklinks,colorlinks,allcolors=cvprblue]{hyperref}

\def\paperID{816} 
\def\confName{CVPR}
\def\confYear{2025}

\title{\includegraphics[scale=0.18]{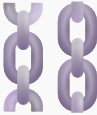} Chain-of-Restoration: Multi-Task Image Restoration Models are Zero-Shot Step-by-Step Universal Image Restorers}

\author{Jin Cao\\
Xi’an Jiaotong University\\
{\tt\small 2213315515@stu.xjtu.edu.cn}\\
\and 
Xiangyong Cao\thanks{Coresponding Author} \\
Xi'an Jiaotong University\\
{\tt\small caoxiangyong@mail.xjtu.edu.cn}\\
\and
Deyu Meng\\
Xi'an Jiaotong University\\
{\tt\small dymeng@mail.xjtu.edu.cn} }

\begin{document}
\maketitle

\begin{abstract}\vspace{-0cm}
Despite previous image restoration (IR) methods have often concentrated on isolated degradations, recent research has increasingly focused on addressing composite degradations involving a complex combination of multiple isolated degradations. However, current IR methods for composite degradations require building training data that contain an exponential number of possible degradation combinations, which brings in a significant burden. To alleviate this issue, this paper proposes a new task setting, i.e. Universal Image Restoration (UIR). Specifically, UIR doesn't require training on all the degradation combinations but only on a set of degradation bases and then removing any degradation that these bases can potentially compose in a zero-shot manner. Inspired by the Chain-of-Thought that prompts large language models (LLMs) to address problems step-by-step, we propose Chain-of-Restoration (CoR) mechanism, which instructs models to remove unknown composite degradations step-by-step. By integrating a simple Degradation Discriminator into pre-trained multi-task models, CoR facilitates the process where models remove one degradation basis per step, continuing this process until the image is fully restored from the unknown composite degradation. Extensive experiments show that CoR can significantly improve model performance in removing composite degradations, achieving comparable or better results than those state-of-the-art (SoTA) methods trained on all degradations. The code will be released \textcolor{blue}{\href{https://github.com/toummHus/Chain-of-Restoration}{at this url}}.
\end{abstract}\vspace{-0.5cm}

\section{Introduction}
\begin{figure*}
    \centering
    \includegraphics[width=0.87\linewidth]{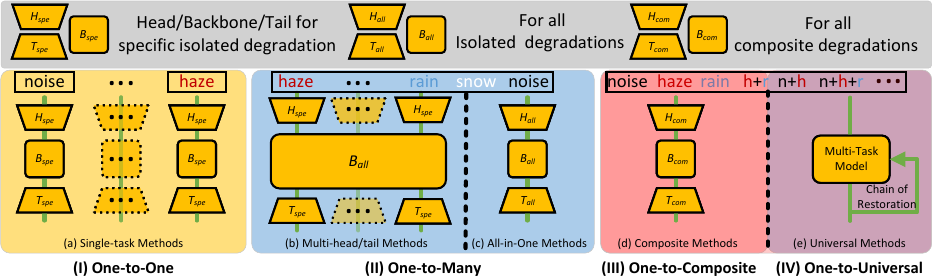}\vspace{-0.2cm}
    \caption{Comparison of task settings and classification of previous image restoration models. \textbf{(I) One-to-One:} In this setting, models are trained on an isolated degradation and tested on it. \textbf{(II) One-to-Many:} In this setting, models are trained on multiple isolated degradations simultaneously and tested on them simultaneously. \textbf{(III) One-to-Composite:} In this setting, models are trained on multiple isolated degradations and composite degradations simultaneously and tested on them simultaneously. \textbf{(IV) One-to-Universal:} In this setting, models are trained on a set of base degradations simultaneously and tested on combinations of these base degradations.}
    \label{fig:comparison}
\end{figure*}

Image restoration plays a crucial role in retrieving high-fidelity imagery from corrupted sources and is extensively applied across various fields, including autonomous navigation, medical imaging, and surveillance systems. Considerable progress has been made in addressing isolated degradations by a One-to-One single-task model (\cref{fig:comparison}(a)) \cite{liang2021swinir,magid2021dynamic, chen2022cross,chen2022simple, zamir2021multi, zamir2022restormer,wang2022uformer,chen2023activating,chen2023dual,guo2024mambair}, such as low-light conditions~\cite{zhou2023low, xu2023low, guo2023low,xu2022snr,liu2021retinex}, haze~\cite{song2023vision, hoang2023transer, zheng2023curricular,qu2019enhanced,guo2022image}, rain~\cite{luo2023local, fu2023continual, du2023dsdnet,jiang2020multi,chen2023learning}, noise~\cite{dabov2007color,tian2020brdnet,zhang2017beyond,zhang2018ffdnet,lehtinen2018noise2noise} and snow~\cite{quan2023image, chen2023lightweight, chen2023msp,chen2022learning,wang2019spatial}. Despite their remarkable achievements in these specific scenarios, real-world conditions often involve unpredictable and variable degradations, posing significant challenges to single-task methods~\cite{li2020all}. 

To overcome the limitations of One-to-One methods, there is an increasing need for universal One-to-Many image restoration techniques that can efficiently handle a variety of degradations within a unified and adaptable framework. Initially, approaches with multiple heads or tails have been introduced~\cite{li2020all, chen2021pre, han2022blind, wang2023smartassign}. These techniques equip distinct heads or tails for each specific degradation type, leveraging a common backbone for processing all degradations, as shown in \cref{fig:comparison}(b). Nonetheless, they require extra parameters for each degradation task and rely on prior knowledge of the degradation type to select the appropriate head or tail, which may not always be feasible in practice. Consequently, recent research has shifted towards All-in-One image restoration methods~\cite{chen2022learning, wang2023context, valanarasu2022transweather, li2022all, ozdenizci2023restoring, potlapalli2024promptir, hair}, employing a single-blind model to manage all the degradations, as depicted in \cref{fig:comparison}(c). Although these methods have accomplished blind image restoration, they are primarily designed for isolated degradation scenarios and thus are not well-suited for composite degradations involving multiple types of degradation.

\begin{figure}[h]
    \centering
    \includegraphics[width=0.85\linewidth]{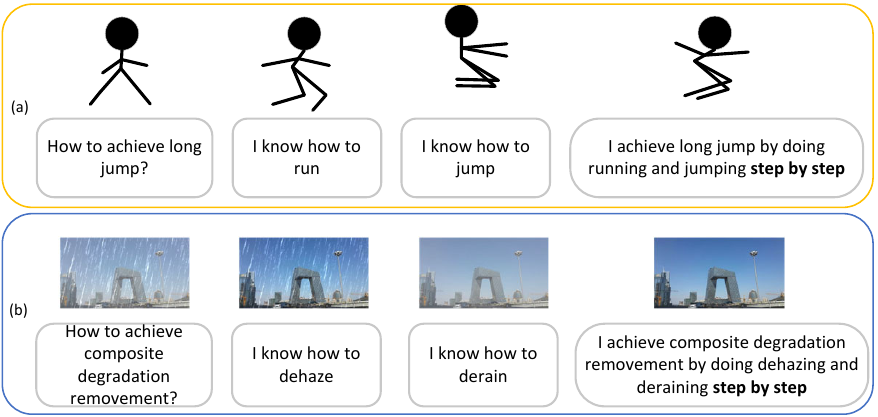}
    \caption{(a) To master long jump, the only requirements are knowing how to run and jump, executed step by step. (b) For an image degraded by both rain and haze, restoring it requires only a model that can dehaze and derain, without any additional training.}
    \label{fig:motivation_jump}
\end{figure}

Recently, OneRestore~\cite{guo2024onerestore} is proposed to address the challenge of composite degradations, as illustrated in \cref{fig:comparison}(d). OneRestore notably accomplishes the removal of composite degradations in a blind or controllable manner by leveraging visual and textual embeddings. However, it has a significant limitation, i.e. it requires training data that encompasses all possible degradations. Given that $n$ isolated degradations can result in $2^n - 1$ distinct degradations, training on the full spectrum becomes increasingly impractical and unreliable as $n$ grows, due to the substantial training costs and the constraints of model capacity.

In numerous practical scenarios, complexity is often a result of interacting with simpler elements. Taking the long jump as an example, it primarily involves running and jumping. Similarly, we suspect that managing complex image degradations could be broken down into handling their individual component, thus multi-task models that are trained on several base degradations should be able to clear complex degradations that are constituted by these components without further training, as depicted in \cref{fig:motivation_jump}. Holding this viewpoint, we first define a new image restoration task setting, named \textbf{Universal Image Restoration} (UIR). In UIR, the model is trained exclusively on a set of degradation bases and tested on both isolated and composite degradations in a zero-shot manner, as shown in \cref{fig:comparison}(e). In this way, we even only need to train the model on $n$ isolated degradations instead of the whole $2^n-1$ degradations.

Subsequently, we try to feed degraded images that exhibit composite degradations into pre-trained multi-task image restoration models. The intriguing discovery is that nearly all these multi-task models predominantly tackle only one degradation at a time when faced with multiple degradations, as validated in \cref{fig:visual_cor}. Leveraging this property and drawing inspiration from Chain-of-Thought (CoT) \cite{cot,zerocot}, which enables Large Language Models (LLMs) to tackle problems step-by-step, we introduce the Chain of Restoration (CoR). This straightforward yet effective method involves integrating a Degradation Discriminator into a pre-trained multi-task model to identify the degradation status of the input image. Consequently, we can apply the model iteratively in a step-by-step manner, with each step dedicated to removing one specific degradation, ultimately restoring the image progressively, shown in \cref{fig:motivation_cot}. By integrating CoR with existing multi-task image restoration methods, we can successfully obtain a Universal Image Restoration (UIR) model. To further facilitate research in this domain, we construct the first dataset (\ie UIRD-12) specifically for UIR. 

In summary, our contributions are three-fold:
\begin{itemize}
    \item We propose a novel task setting for image restoration, \ie Universal Image Restoration (UIR). In this framework, models are restricted to training on a set of degradation bases and are challenged to handle both isolated and composite degradations combined by these bases. Then we present the first dataset designed for UIR, \ie Universal Image Restoration Dataset (UIRD-12).
    \item We first propose a simple and effective method, \ie Chain of Restoration (CoR), to build the UIR model. By incorporating a simple Degradation Discriminator, CoR transforms a multi-task model into a UIR model in a zero-shot manner, requiring no additional training.
    \item Our comprehensive experiments demonstrate the robust capability of CoR. It notably enhances the performance of multi-task models on UIR tasks, often matching or even surpassing SoTA models trained on all the degradations.
\end{itemize}

\section{Related Works}
\par

\quad \textbf{Image Restoration for Composite Degradations.} 
While One-to-One and One-to-Many models address isolated degradations, recent research has shifted focus towards composite degradation removal~\cite{Zhou2022TaskAN,guo2024onerestore,duan2024uniprocessor,restoreagent}. Notably, OneRestore \cite{guo2024onerestore} pioneers a unified model leveraging visual and textual embeddings to confront the challenge of composite degradation removal. Nonetheless, it requires comprehensive training data across all degradation types, resulting in high costs and being limited by the model's capacity. Uniprocessor \cite{duan2024uniprocessor} and RestoreAgent \cite{restoreagent} both identify step-by-step removal of composite degradation as feasible for their specialized models. However, \cite{duan2024uniprocessor} only mentions this without further investigation. Moreover, \cite{restoreagent} requires individual model training for each type and intensity of degradation, followed by extra fine-tuning of a large vision-text model for every possible sequence permutation of isolated degradations, rendering the approach prohibitively expensive and impractical. In contrast, our CoR, with a single pre-trained multi-task model, necessitates merely a simple and cheap Degradation Discriminator for step-by-step composite degradation removal, proving its simplicity and efficacy, and demonstrating that any single multi-task model possesses the generalization capability to remove composite degradations in a zero-shot manner. Considering the constraints of previous task settings involving composite degradation, we also introduce Universal Image Restoration (UIR). In this task setting, models are trained on a set of degradation bases and evaluated on their ability to remove both isolated and composite degradations.

\textbf{Zero-Shot Learning} 
Zero-shot learning is a paradigm in which machine learning models perform tasks on new data without prior training on similar examples~\cite{review_zsl}. The rise of Large Language Models (LLMs) and prompting techniques~\cite{bert,brown2020language,liu2021pre,liu2021makes,ouyang2022training} has intensified interest in zero-shot learning within the LLM domain~\cite{zerocot,zhang2023automatic,hou2024largelanguagemodelszeroshot,NEURIPS2023_3eb7ca52}. Specifically, Zero-Shot CoT~\cite{zerocot} shows that guiding LLMs to reason step by step significantly improves their performance on complex problems. Motivated by this, we view composite degradations as a sum of several base degradations and propose an iterative method to sequentially remove each type of degradation, progressively restoring the image.

\begin{algorithm}[t]
\caption{Pseudocode of CoR in a PyTorch-like style.}
\label{alg:cor}
\definecolor{codeblue}{rgb}{0.25,0.5,0.5}
\lstset{
  backgroundcolor=\color{white},
  basicstyle=\fontsize{6pt}{6pt}\ttfamily\selectfont,
  columns=fullflexible,
  breaklines=true,
  captionpos=b,
  commentstyle=\fontsize{7.2pt}{7.2pt}\color{codeblue},
  keywordstyle=\fontsize{7.2pt}{7.2pt},
}
\begin{lstlisting}[language=python]
# X: input image with unknown composite degradation
# M: Multi-task model trained on degradation bases 
# cls: The image classifier to discriminate degradations
# ep_o: Soft margin of order, a positive float
# orders:  The orders of corresponding bases, a list of integers
# ep_bs: The soft margins of corresponding bases, a list of floats
# n: the number of bases

def Degradation_Discriminator(X,cls,ep_o,orders,ep_b):
    v=cls(X)
    for idx in range(len(orders)):
        # up date v via Eq.(6)
        v[idx]+=ep_o*orders[idx]+ep_bs[idx] 
    return v.argmax(dim=-1)

def Chain_of_Restoration(X,M,cls,n,ep_o,orders,ep_b):
    type = Degradation_Discriminator(X,cls,ep_o,orders,ep_b) # the degradation situation of I
    while type != n+1 : # the image is not clean yet
        X = M(X,type) # a single step
        type = Degradation_Discriminator(X,cls,ep_o,orders,ep_b)  # update the degradation situation
    return X # Restored image

\end{lstlisting}
\end{algorithm}

\section{Method}\label{sec:method}

\subsection{Chain of Restoration}
As previously introduced, we observed that multi-task models trained on multiple degradations typically address only one degradation when faced with an image containing composite degradations. Our proposed Chain of Restoration (CoR) capitalizes on this behavior. For an input image $\mathbf{X}_0$ with composite degradation comprising $T$ components and a multi-task model $\mathbf{M}$ trained on these components, CoR operates step-by-step as follows:
\begin{equation}
    \begin{aligned}
        & \mathbf{type} = DD (\mathbf{X}_{i-1}),\\
        & \mathbf{X}_{i} = \mathbf{M}(\mathbf{X}_{i-1}, \mathbf{type}), \quad i = 1, 2, \cdots, T
         \label{eq:cor_iter}
    \end{aligned}
\end{equation}
Here, $\mathbf{type}$ is the degradation type identified from $\mathbf{X}_{i-1}$ by the $DD$ (Degradation Discriminator). Ideally, the model removes one degradation per step, and after $T$ steps, outputs a clear restored image $\mathbf{X}_{T}$ with the degradations removed.

\subsection{Degradation Basis}
The essence of CoR is the progressive elimination of composite degradations by addressing their individual components. Models are trained to remove these elemental components, and each of these components is defined as a degradation basis, which essentially refers to an individual degradation. An $n$-order degradation refers to a scenario where $n$ isolated degradations are present simultaneously. For example, "rain" and "haze" are termed 1-order bases, while combinations such as "rain+haze" and "haze+snow" are referred to as 2-order bases. Accordingly, we define $n$-order models as those trained on all the given degradation bases whose order $\leq n$.

To effectively apply CoR to remove a specific degradation $d = s_1 + s_2 + \cdots + s_m$, where each $s_i$ represents an isolated degradation, it is crucial that the model has been trained on bases $b_1, b_2, \ldots, b_n$ that can be directly combined to form $d$. Given degradation bases, the combination of degradation bases is defined as:
\begin{equation}
    combine(b_1, b_2) = b_1 + \text{"+"} + b_2,
\end{equation}
where $b_1$, $b_2$, and "+" are treated as strings, and $+$ denotes string concatenation. For instance, $$combine(\text{haze+snow}, \text{noise}) = \text{haze+snow+noise}.$$ We define the equality of two degradations $d_1$ and $d_2$ if one is a permutation of the other:
\begin{equation}
    d_1 = d_2 \Leftrightarrow d_1 \in \textit{Permu}(d_2) \Leftrightarrow d_2 \in \textit{Permu}(d_1),
\end{equation}
where for a degradation $d = s_1 + s_2 + \cdots + s_n$, the permutation set $\textit{Permu}(d)$ is:
\begin{equation}
    \begin{aligned}
        \textit{Permu}(d) = \{s_{k_1} + s_{k_2} + \cdots + s_{k_n} \mid k_1, k_2, \ldots, k_n \\ \text{ is a permutation of } 1, 2, \ldots, n\}.
    \end{aligned}
\end{equation}
Determining how multiple bases $b_1, b_2, \ldots$ can combine into a degradation $d$ while maximizing the order of the bases, is computationally complex (NP-hard). Moreover, the type of degradation in an input image is often unknown, thus requiring a Degradation Discriminator to identify it.

\subsection{Degradation Discriminator}\label{sec:method:DD}
A Degradation Discriminator (DD) is essential to determine an input image's degradation state. Specifically, with a multi-task image restoration model pre-trained on degradation bases $b_1, b_2, \ldots, b_n$, the DD operates in two scenarios:
 \par
\textbf{Scenario 1: the given model is blind.} In the case of a blind multi-task image restoration model (e.g. All-in-One models), DD serves only to identify if the input image is clean or degraded. The DD is trained on clean images as well as those with any degradation among $b_1, b_2, \ldots, b_n$, effectively acting as a binary classifier. Specifically, if DD outputs "clean", the iterative process of CoR stops and outputs the restored image; otherwise, it continues.
\par
\textbf{Scenario 2: the given model is non-blind.} For non-blind multi-task image restoration models (e.g. multi-head/tail methods), the DD must recognize the specific type of degradation present. Thus DD acts as a multiple classifier, designed to differentiate between clean images and those with one of degradations bases $b_1, b_2, \ldots, b_n$. However, there exist two challenges: (I) The identified degradation type may not be unique, e.g., an image with "haze+rain" could correctly be labelled as "rain," "haze," or "haze+rain." We aim to address higher-order degradations first to optimize performance. (II) The restoration sequence of the bases is not predetermined or controllable, which can impact the outcome. To address these issues, pre-defined hyperparameters soft margins $\epsilon_o$ and $\epsilon_{b_i}$ are introduced. Given an input image $\mathbf{X}$ with an unknown degradation, the DD generates a probabilistic vector $\mathbf{v} \in \mathbb{R}^{n+1}$ as
\begin{equation}
    \mathbf{v} = Softmax(DD(\mathbf{X})).
\end{equation}
The $i$-th element $\mathbf{v}_i$ represents the probability of selecting the degradation basis $b_i$ in this step, while the $(n+1)$-th element corresponds to the probability of the image being in a "clean" state. Given that the order of $b_i$ is $o_i$, we can derive the revised probabilistic vector $\mathbf{v}' \in \mathbb{R}^n$ as:
\begin{equation}
    \mathbf{v}_i'=\mathbf{v}_i+o_i*\epsilon_o+\epsilon_{b_i}. \label{eq:update_v}
\end{equation}
Here, $\epsilon_o$ represents the soft margin for degradation order, a positive value that favours the selection of higher-order bases. Meanwhile, $\epsilon_{b_i}$ is the soft margin for the degradation basis $b_i$, used to give preference to specific degradation bases. The final choice of basis $b_t$ is determined by:
\begin{equation}
    t=\mathop{\arg\max}_{t}\mathbf{v}_t'.
\end{equation}
The pseudocode of the CoR is provided in \cref{alg:cor}

\subsection{Method Complexity}\label{method:complexity}
Given $n$ isolated degradations $s_1, s_2, \ldots, s_n$ and their corresponding combined degradations $\{s_{k_1} + s_{k_2} + \cdots + s_{k_m} \mid 1 \leq k_1 < k_2 < \cdots < k_m \leq n, 1 \leq m \leq n\}$, we first define $\phi_n(k)$ as follows:
\begin{equation}
    \phi_n(k) = \sum_{t=1}^k C_n^t, \text{ where } C_n^t = \frac{n!}{t!(n-t)!}, 1 \leq k \leq n.
\end{equation}
Here, $\phi_n(k)$ describes the number of bases that a $k$-order model (trained on all given degradations of order $\leq k$) will be trained on, given $n$ isolated degradations. Assuming each degradation has $N$ training pairs, a $k$-order model will be trained on $\phi_n(k) \times N$ images for each epoch. We then define the training ratio $TR_n(k)$ as:
\begin{equation}
    TR_n(k) = \frac{\phi_n(k)}{\phi_n(1)} = \frac{\phi_n(k)}{n}, \quad 1 \leq k \leq n. \label{eq:tr}
\end{equation}
Here, $TR_n(k)$ indicates the factor by which the training time for a $k$-order model exceeds that for a corresponding $1$-order model, per training epoch. Although $TR_n(k)$ can't simplify to a concise formula, we know the following: (1) $TR_n(1) = 1$. (2) $TR_n(n) = \frac{2^n - 1}{n}$. (3) $TR_n(k+1) > TR_n(k)$. 

Next, we consider the inference time. We start by defining $\varphi_n(k)$ as follows:
\begin{equation}
    \varphi_n(k) = \sum_{t=1}^n C_n^t \left\lceil \frac{t}{k} \right\rceil, \quad 1 \leq k \leq n,
\end{equation}
where $\left\lceil * \right\rceil$ denotes the ceiling function. Similarly, we define the inference ratio $IR_n(k)$ as follows:
\begin{equation}
    IR_n(k) = \frac{\varphi_n(k)}{\varphi_n(n)} = \frac{\varphi_n(k)}{2^n - 1}, \quad 1 \leq k \leq n. \label{eq:ir}
\end{equation}
Assuming all degradations have the same probability of occurrence and the model always prioritizes bases with higher orders, on average, the inference time of a $k$-order model is $IR_n(k)$ times that of an $n$-order model (i.e., the end-to-end model like \cite{guo2024onerestore}). Although $IR_n(k)$ does not have a simple expression, we obtain that: (1) $IR_n(1) = \frac{n2^{n-1}}{2^n - 1} \approx \frac{n}{2}$. (2) $IR_n(n) = 1$. (3) $IR_n(k+1) < IR_n(k)$.
\par
\begin{figure}[h]
      \centering
    \begin{subfigure}{0.5\linewidth}
        \centering
        \includegraphics[width=0.99\linewidth]{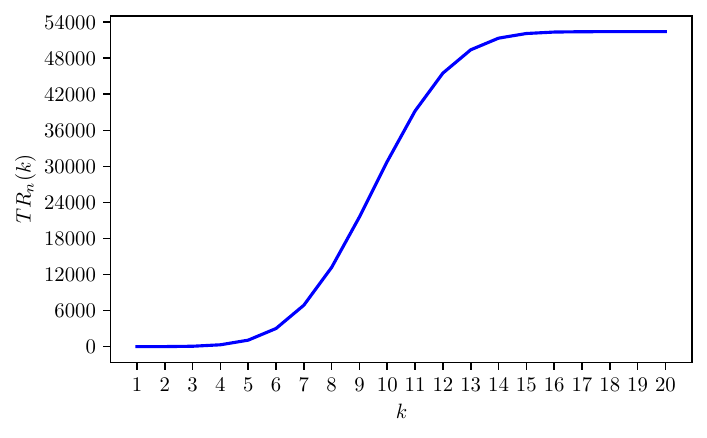}
        \caption{$TR_n(k)$}
        \label{fig:complexity_TR}
    \end{subfigure}%
    \hfill
    \begin{subfigure}{0.5\linewidth}
        \centering
        \includegraphics[width=0.94\linewidth]{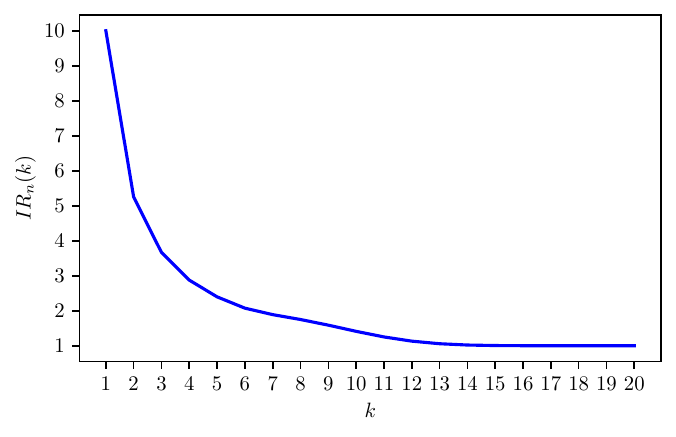}
        \caption{$IR_n(k)$}
        \label{fig:complexity_IR}
    \end{subfigure}%

    \vspace{-0cm}\caption{Visualization of $TR_n(k)$ and $IR_n(k)$ when $n=20$.}\vspace{-0cm}
        \label{fig:complexity}
\end{figure}

 
\textbf{Analysis.} \cref{eq:ir} and \cref{eq:tr} show that as the order $k$ of the model increases, it tends to gain more training time and less inference time, as shown in \cref{fig:complexity}. It can be seen that when $k$ increases, $TR_n(k)$ can grow at an exponential rate and $IR_n(k)$ decreases from fast to slow. Still, the addition of inference time when $k=1$ is acceptable and the addition of training time when $k=n$ is not acceptable. Furthermore, a higher order doesn't mean better performance since the capacity of the models is always limited; training degradations too much can impair the model's overall performance. Considering these, it's recommended to use a relatively low-order model when using CoR.

\section{Experiment} \label{sec:exp}
\subsection{Experiment Settings}
\textbf{Implementation Details.} Most results and pre-trained models in this paper are directly from previous works~\cite{li2022all,potlapalli2024promptir,hair,guo2024onerestore}. Most of the re-trained models (including the image classifiers) are trained with a batch size of 32 (64 for the classifiers) on 8 NVIDIA GeForce RTX 3090 Ti GPUs. The network optimization is guided by an $L_1$ loss function, employing the AdamW optimizer \cite{loshchilov2017fixing} with parameters $\beta_1 = 0.9$ and $\beta_2 = 0.999$. The learning rate is set to $2e-4$ ($2e-3$ for the classifiers). To enhance the training data, input patches of size 128 $\times$ 128 are utilized, with random horizontal and vertical flips applied to the images to augment the dataset. All the classifiers in this paper are MobileNetV3 small~\cite{Howard_2019_ICCV} and trained using the same dataset as the models.
\par 
\subsubsection{Datasets} 
\quad \textbf{The Synthesised Dataset UIRD-12.} We propose a new dataset for UIR, \ie Universal Image Restoration Dataset-12 (UIRD-12). The training set of UIRD-12 closely follows previous All-in-One works~\cite{li2022all,potlapalli2024promptir}: BSD400~\citep{arbelaez2010contour} and WED~\citep{ma2016waterloo} datasets for training on Gaussian denoising ($\sigma=\{15,25,50$\}); Rain100L dataset~\citep{yang2020learning} for derain; SOTS dataset~\citep{li2018benchmarking} for dehaze. For the test set, we use BSD68~\citep{martin2001database_bsd}, Urban100~\citep{huang2015single}, Rain100L~\citep{yang2020learning}, SOTS~\citep{li2018benchmarking} to synthesise 12 categories of image degradations and their clear counterparts. These degradations include \textit{n1, n2, n5, r, h, h+r, h+n1, h+n5, r+n1, r+n5, h+r+n1, h+r+n5}. (n1: noise($\sigma=15$), n2: noise($\sigma=25$), n5: noise($\sigma=50$), r: rain, h: haze.) Each category contains 100 images. All models are trained on the 5 isolated degradations and tested on the 12 degradations.

\textbf{Datasets Settings.} To better verify the effect of CoR, we utilize not only the UIRD-12 dataset in our experiments but also the CDD-11 dataset~\cite{guo2024onerestore}, which encompasses 11 degradation types including \textit{l}, \textit{h}, \textit{r}, \textit{s}, \textit{l+h}, \textit{l+r}, \textit{l+s}, \textit{h+r}, \textit{h+s}, \textit{l+h+r}, and \textit{l+h+s}. (l: low-light, s: snow, r: rain, h: haze.) Unlike UIRD-12, CDD-11 contains all the corresponding training and testing data for each degradation type. Specifically, 1-order models are trained only on 1-order degradations (\ie \textit{l}, \textit{h}, \textit{r}, \textit{s}), 2-order models are trained on all degradations except \textit{l+h+r} and \textit{l+h+s}, and 3-order models are end-to-end models trained on all degradations.

\textbf{Compared Methods and Evaluation Metrics.}
To validate the effectiveness of our proposed CoR, we experiment it with a range of methods. In the experiment on UIRD-12, we primarily select 4 One-to-Many methods including AirNet \cite{li2022all}, PromptIR \cite{potlapalli2024promptir}, InstructIR \cite{conde2024instructir}, HAIR \cite{hair}, and a One-to-Composite method OneRestore \cite{guo2024onerestore}. We combine these methods with CoR to demonstrate its effectiveness. In the experiment on CDD-11, we compare low-order methods integrated with CoR against end-to-end methods, comprising 9 One-to-One image restoration methods (MIRNet~\cite{zamir2020learning}, MPRNet~\cite{zamir2021multi}, MIRNetv2~\cite{zamir2022learning}, Restormer~\cite{zamir2022restormer}, DGUNet~\cite{mou2022deep}, NAFNet~\cite{chen2022simple}, SRUDC~\cite{song2023under}, Fourmer~\cite{zhou2023fourmer}, OKNet~\cite{cui2024omni}) and 6 One-to-Many image restoration methods (AirNet~\cite{li2022all}, TransWeather~\cite{valanarasu2022transweather}, WeatherDiff~\cite{ozdenizci2023restoring}, PromptIR~\cite{potlapalli2024promptir}, WGWSNet~\cite{zhu2023learning}, HAIR \cite{hair}, and the  One-to-Composite method OneRestore \cite{guo2024onerestore}). Notably, some methods are trained on different orders. Additionally, we employ the Peak Signal-to-Noise Ratio (PSNR) and Structure Similarity Index Measure (SSIM) as our evaluation metrics.

\begin{figure*}
    \centering
    \includegraphics[width=0.8\linewidth]{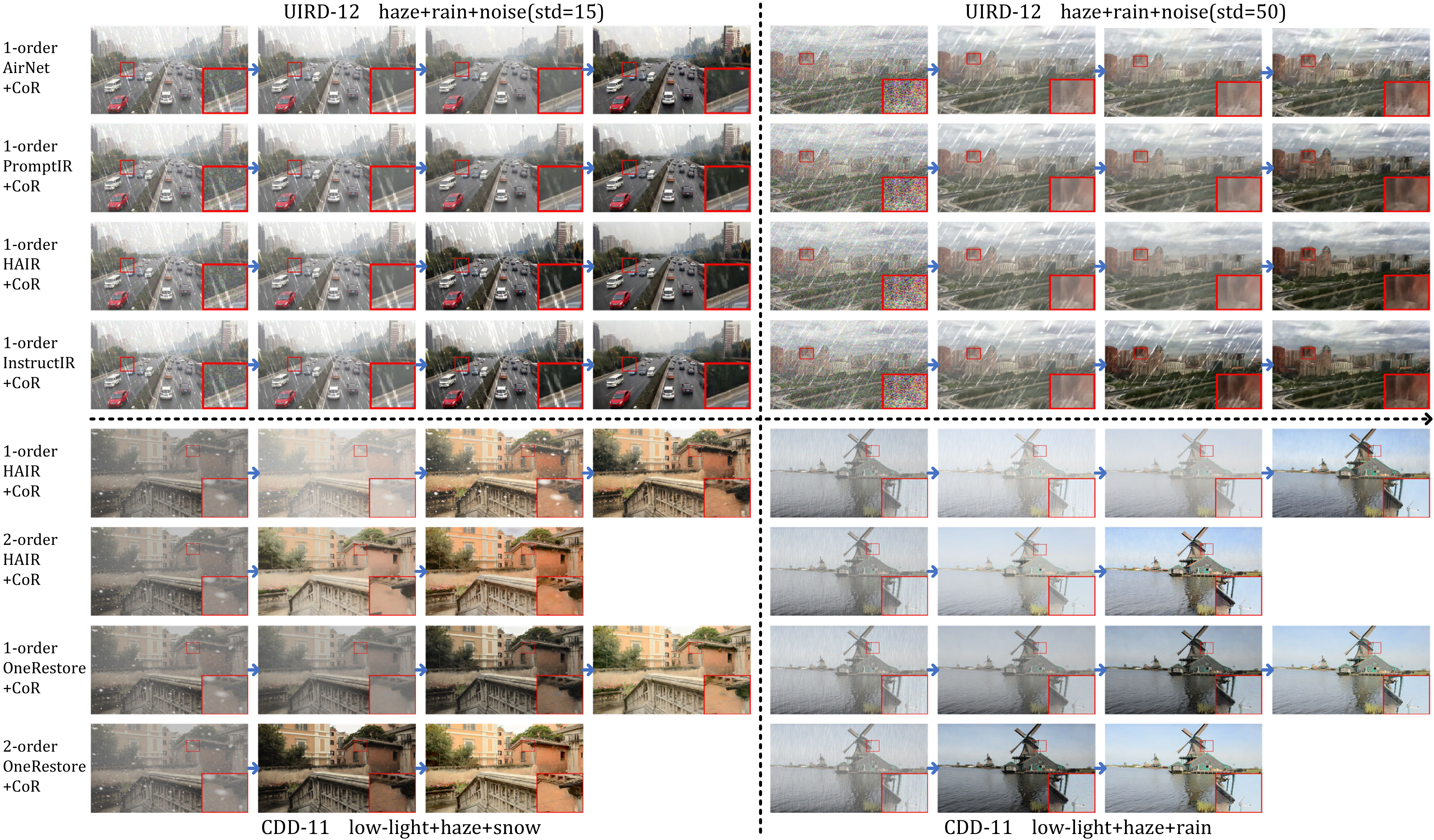}
    \caption{The visualization of the step-by-step process of CoR with different methods on UIRD-12 and CDD-11.}
    \label{fig:visual_cor}
\end{figure*}

\subsection{Results}

\begin{figure}[h]
      \centering
    \begin{subfigure}{0.5\linewidth}
        \centering
        \includegraphics[width=1\linewidth]{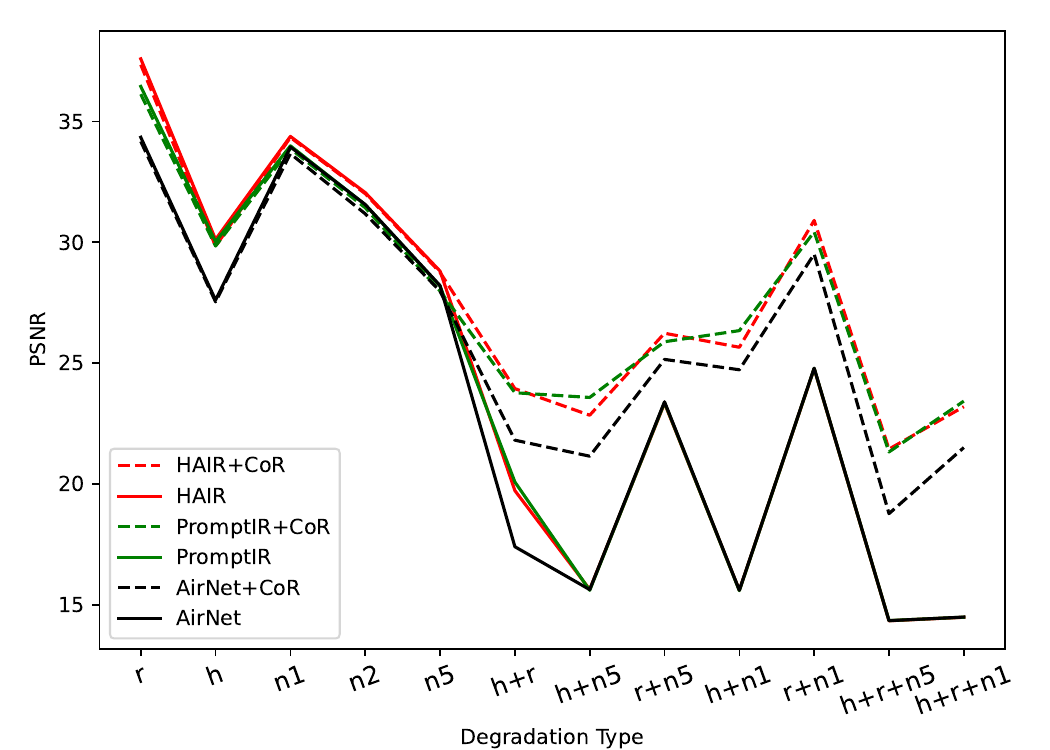}
        \caption{CoR for blind methods.}
        \label{fig:uird_blind_PSNR}
    \end{subfigure}%
    \hfill
    \begin{subfigure}{0.5\linewidth}
        \centering
        \includegraphics[width=1\linewidth]{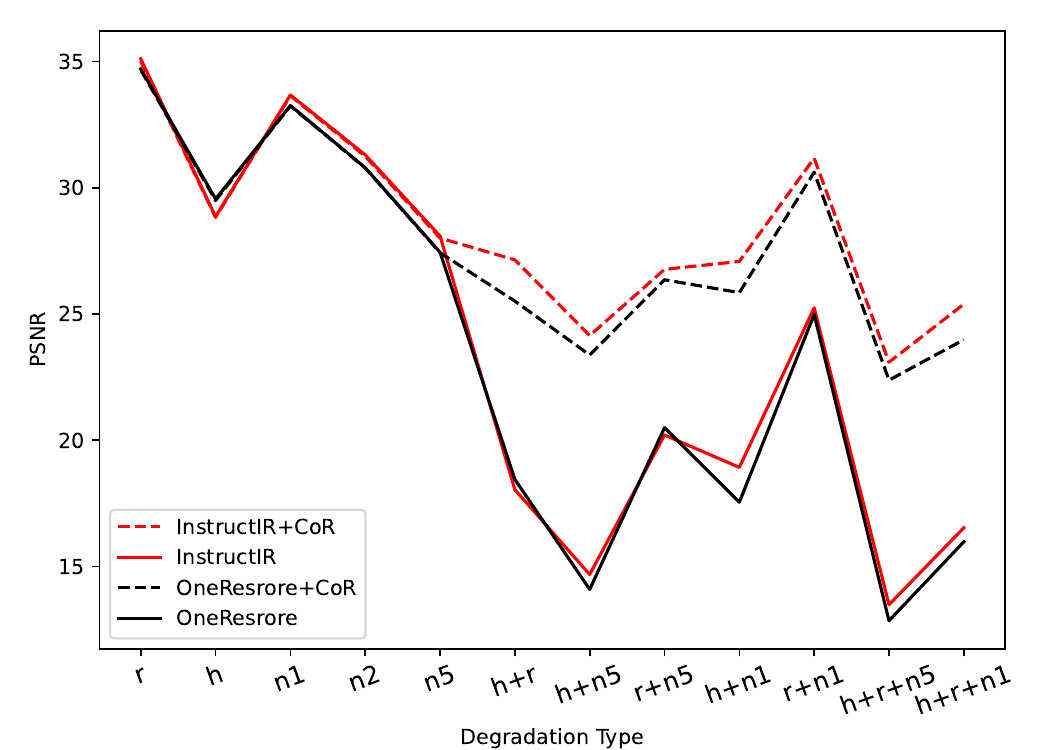}
        \caption{CoR for non-blind methods.}
        \label{fig:uird_nonblind_PSNR}
    \end{subfigure}%

    \vspace{-0cm}\caption{Visual Comparison of Multi-Task Models' Performance with and without CoR on UIRD-12.}\vspace{-0cm}
        \label{fig:uird12_PSNR}
\end{figure}

\textbf{Results on UIRD-12.} The results of all the methods on UIRD-12 are presented in \cref{tab:uird12} and \cref{fig:uird12_PSNR}. It's obvious that our proposed CoR can significantly improve the performance of these 1-order multi-task models on composite degradation removal with only the addition of a simple Degradation Discriminator, while having nearly no impact on isolated degradation removal. What's more, we find that non-blind models gain obviously more increment from CoR than blind models. This is because non-blind methods know the exact degradation type they are processing, thus alleviating the Degradation Coupling issue described in \cref{sec:limitation}. The results show the effectiveness and feasibility of CoR.

\begin{table}[h]
        \scriptsize        
        \setlength{\tabcolsep}{1pt}

        \centering
        \caption{Average quantitative performance comparison on the UIRD-12 dataset. "+CoR" indicates the direct integration of the pre-trained model with CoR, while "+1.52M" signifies that CoR introduces only an additional 1.52 M parameters.}
        \label{tab:uird12}
        \begin{tabular}{c|l|c|c|c|c|c}
            \toprule
            Types & Methods     &Order      &  Blind &PSNR $\uparrow$& SSIM $\uparrow$                                 & \#Params \\ \midrule
           - & Input    &   -    &   -    & 17.19 & 0.4523  & - \\ \midrule
            \multirow{4}{*}{\centering One-to-Many} & AirNet~\cite{li2022all}        &    1      &  \cmark& 23.44 & 0.7902                              & 8.93M    \\

                        \multirow{5}{*}{} & PromptIR~\cite{potlapalli2024promptir} &   1   & \cmark& 24.03 & 0.7904     & 35.59M   \\
           
            & InstructIR~\cite{conde2024instructir}          &    1    &  \xmark & 23.68 & 0.7042                               & 15.94M   \\
            & HAIR~\cite{hair}          &    1    &  \cmark & 24.23 & 0.7939                               & 28.56M   \\
       
            \midrule
            \multirow{1}{*}{One-to-Composite} & OneRestore~\cite{guo2024onerestore}    & 1  &     \xmark      & 23.34&0.6952     & 5.98M    \\ 
                \midrule

             \multirow{5}{*}{\centering One-to-Universal} & AirNet\cite{li2022all}+CoR        &    1      &  \cmark& 26.43 & 0.8407                               &  \multirow{5}{*}{+1.52M}   \\
            \multirow{5}{*}{} & PromptIR\cite{potlapalli2024promptir}+CoR &   1   & \cmark& 27.83 & 0.8556  &    \\               
            & InstructIR\cite{conde2024instructir} +CoR         &    1    &  \cmark & 28.44 & 0.8726                               &    \\
             & HAIR\cite{hair}+CoR     &    1    &  \cmark & 28.04 & 0.8664     &    \\
              & OneRestore\cite{guo2024onerestore}+CoR     & 1  &     \cmark      & 27.78&0.8607    &     \\ 

            \bottomrule
        \end{tabular}
    \end{table}

\par
\begin{figure}
    \centering
    \includegraphics[width=0.7\linewidth]{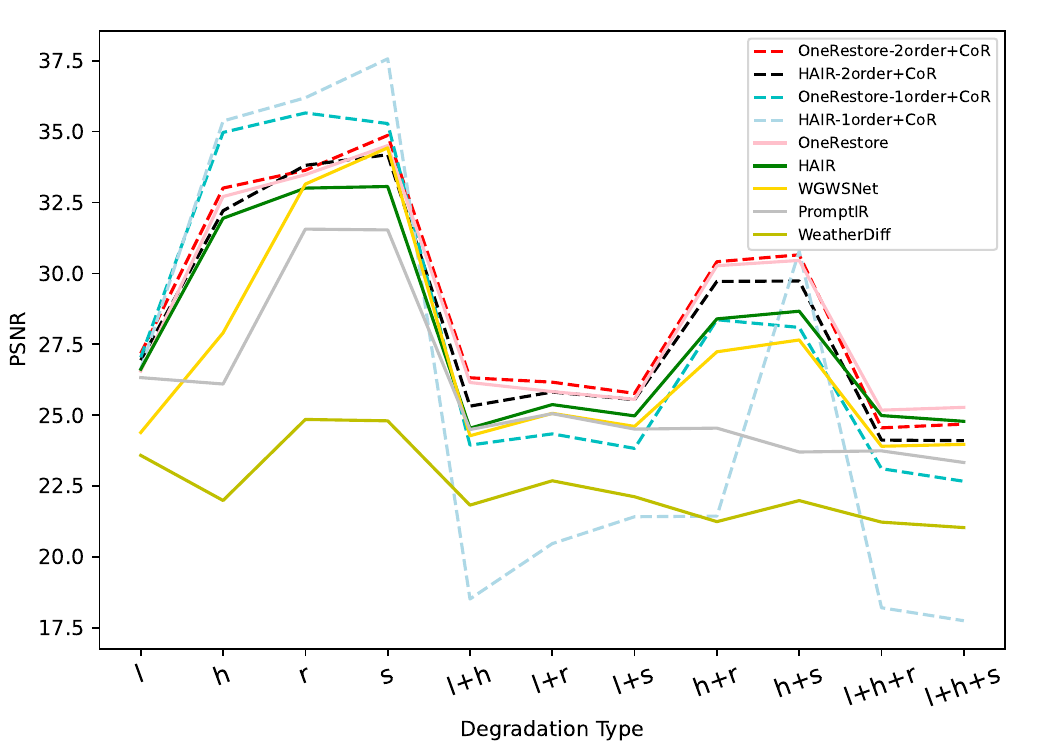}
    \caption{Visualization of performance of multi-task models on CDD-12. Methods without "order" are all 3-order end-to-end models trained on all degradations.}
    \label{fig:cdd11_psnr}
\end{figure}

\textbf{Results on CDD-11.} The results of all the methods on CDD-11 are presented in \cref{tab:cdd11} and \cref{fig:cdd11_psnr}. It can be seen that with CoR, models with low orders can achieve comparable or even superior performance to models trained on all degradations. Notably, the 2-order OneRestore with CoR achieves comparable performance with the 3-order OneRestore, and the 2-order HAIR with CoR surpasses the 3-order HAIR in both PSNR and SSIM, demonstrating the effectiveness of CoR. As previously discussed, due to limited capacity, training the model on all degradations can lead to decreased performance on each degradation, which is why 2-order methods with CoR can perform better with less training. However, as shown in \cref{fig:cdd11_psnr}, 1-order HAIR with CoR fail to achieve satisfactory results in composite degradations with low-light due to the Degradation Coupling described in \cref{sec:limitation}. Additionally, we provide a training time comparison in \cref{tab:train_time}. It is evident that models with lower orders require less time and thus have lower training costs to converge, not only due to the reduced time cost per epoch but also because less data requires fewer epochs to converge.

\begin{table}
        \scriptsize
        \setlength{\tabcolsep}{1pt}

        \centering
        \caption{Comparison of average quantitative results on the CDD-11 dataset. "+CoR" indicates the direct integration of the pre-trained model with CoR. \textcolor{red}{Red} and \textcolor{blue}{blue} represent the best and second-best results, respectively.}
        \label{tab:cdd11}
        \begin{tabular}{c|l|c|c|c|c|c}
            \toprule
            Types & Methods     &Order      &  Blind &PSNR $\uparrow$& SSIM $\uparrow$                                 & \#Params \\ \midrule
           - & Input                                   &   -    &      -       & 16.00 & 0.6008                               &      -   \\ \midrule
            \multirow{9}{*}{One-to-One} & MIRNet  ~\cite{zamir2020learning}      &    3     &  \cmark & 25.97 & 0.8474                               & 31.79M   \\
            \multirow{9}{*}{} & MPRNet~\cite{zamir2021multi}         &     3     & \cmark & 25.47 & 0.8555                               & 15.74M   \\
            \multirow{9}{*}{} & MIRNetv2~\cite{zamir2022learning}           &  3 & \cmark &25.37 & 0.8335                               & 5.86M    \\
            \multirow{9}{*}{} & Restormer~\cite{zamir2022restormer}   &     3    &\cmark & 26.99 & {0.8646}                 & 26.13M   \\
            \multirow{9}{*}{} & DGUNet~\cite{mou2022deep}             &    3     & \cmark& 26.92 & 0.8559                               & 17.33M   \\
            \multirow{9}{*}{} & NAFNet~\cite{chen2022simple}        &      3     & \cmark& 24.13 & 0.7964                               & 17.11M   \\
            \multirow{9}{*}{} & SRUDC~\cite{song2023under}           &     3     & \cmark&{27.64} & 0.8600                 & 6.80M    \\
            \multirow{9}{*}{} & Fourmer~\cite{zhou2023fourmer}     &        3            &\cmark &23.44 & 0.7885                 &  0.55M   \\ 
            \multirow{9}{*}{} & OKNet~\cite{cui2024omni}          &       3    &  \cmark & 26.33 & 0.8605                 &   4.72M  \\ 
            \midrule
            \multirow{8}{*}{\centering One-to-Many} & AirNet~\cite{li2022all}        &    3      &  \cmark& 23.75 & 0.8140                               & 8.93M    \\
            \multirow{4}{*}{} & TransWeather~\cite{valanarasu2022transweather} &  3   &\cmark &23.13 & 0.7810                               & 21.90M   \\
            \multirow{4}{*}{} & WeatherDiff~\cite{ozdenizci2023restoring}   &   3  &\cmark &22.49 & 0.7985                                        & 82.96M   \\
            \multirow{5}{*}{} & PromptIR~\cite{potlapalli2024promptir} &   3   & \cmark& 25.90 & 0.8499                                        & 38.45M   \\
            \multirow{4}{*}{} & WGWSNet~\cite{zhu2023learning}          &    3    &  \xmark & 26.96 & 0.8626                               & 25.76M   \\

              & HAIR~\cite{hair}          &    1    &  \cmark & 23.01 & 0.7632                               & 28.56M   \\
              & HAIR~\cite{hair}          &    2    &  \cmark & 27.65 & 0.8655                               & 28.56M   \\
            & HAIR~\cite{hair}          &    3    &  \cmark & 27.85 & 0.8663                               & 28.56M   \\
              \midrule
            \multirow{3}{*}{One-to-Composite} 
            & OneRestore~\cite{guo2024onerestore}    & 1  &     \xmark      & {21.43}&{0.7226}     & 5.98M    \\ 
            & OneRestore~\cite{guo2024onerestore}    & 2  &     \xmark      & {27.28}&{0.8437}     & 5.98M    \\ 
            & OneRestore~\cite{guo2024onerestore}    & 3  &     \xmark      & {\color{blue}28.72}&{\color{red}0.8821}     & 5.98M    \\ 
            \midrule
            \multirow{4}{*}{One-to-Universal} & HAIR\cite{hair}+CoR     &    1    &  \cmark & 25.85 & 0.8289     &  \multirow{4}{*}{+1.52M}  \\
              & HAIR\cite{hair}+CoR          &    2    &  \cmark & 28.33 & 0.8688                               &    \\
              & OneRestore\cite{guo2024onerestore}+CoR     & 1  &     \cmark      & {27.94}&{0.8541}     &    \\ 
            & OneRestore\cite{guo2024onerestore}+CoR     & 2  &     \cmark      & {\color{red}28.84}&{\color{blue}0.8794}     &    \\
            
            \bottomrule
        \end{tabular}
        \vspace{-1.9mm}
    \end{table}

\begin{table}
    \centering
        \centering
        \scriptsize
        \fboxsep0.1pt
        \setlength\tabcolsep{3.3pt}
        \captionof{table}{{Comprison of training time.} Results are from CDD-11 on 8 NVIDIA GeForce RTX 3090 Ti GPUs, we re-train all the methods until convergence (including the training of Classifier). If the average training loss remains nearly unchanged over five consecutive epochs, we consider the model to have converged.}\vspace{-0cm}
        \label{tab:train_time}
    \begin{tabular}{cccccccc}
            \toprule
                Method &  HAIR & HAIR  & HAIR  & OneRestore & OneRestore & OneRestore \\
             \midrule

            Order & 1 & 2 & 3  & 1 & 2 & 3 \\
            \midrule
            Time & 30.71h & 76.25h & 109.76h & 21.39h & 53.62h& 71.70h& \\
             \bottomrule
        \end{tabular}\vspace{-0.3cm}
   
\end{table}

\par
\begin{figure*}
    \centering
    \includegraphics[width=0.84\linewidth]{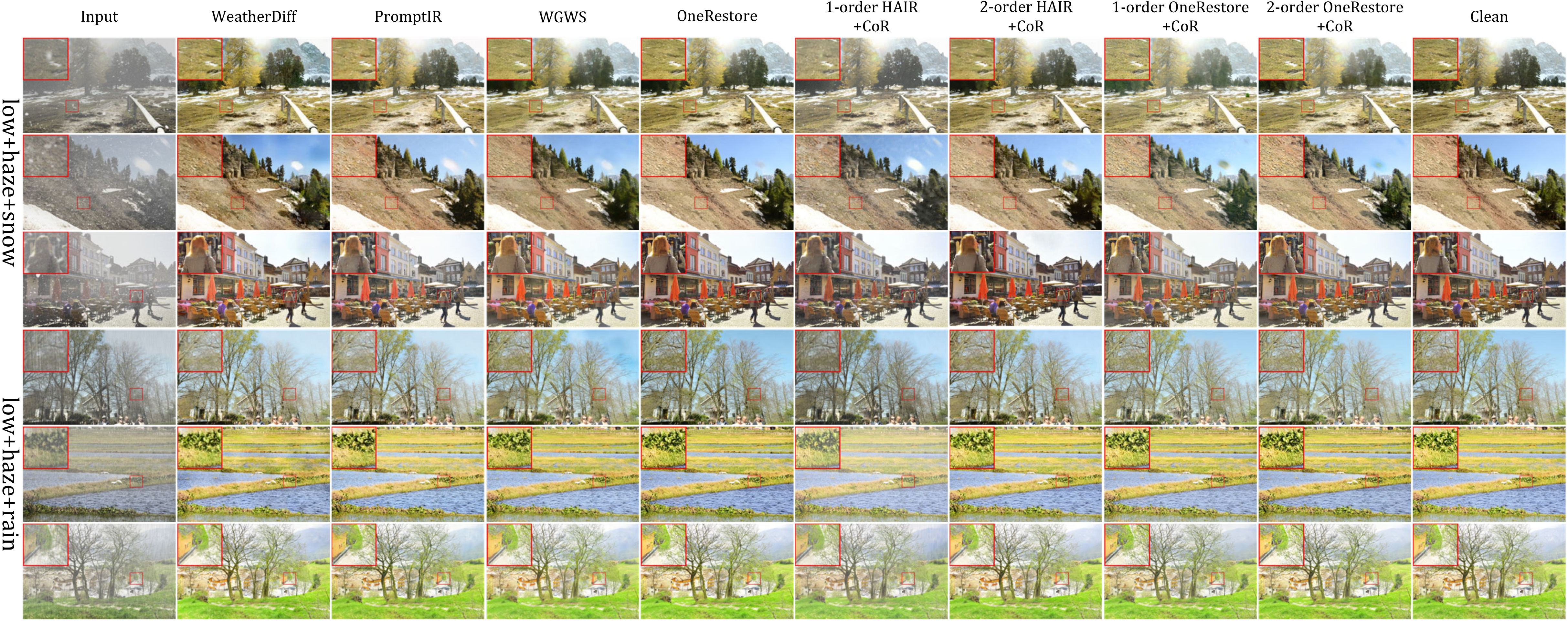}
    \caption{Visual comprison on CDD-11. Zoom for better visual effects.}\vspace{-0.3cm}
    \label{fig:visual_cdd}
\end{figure*}

 
\textbf{Visual Results.} We present visual results in \cref{fig:visual_cor} and \cref{fig:visual_cdd}. \cref{fig:visual_cor} illustrates how CoR assists multi-task models in step-by-step removal of composite degradations. It is evident that each step typically addresses only one degradation basis that the models are trained on, which is a common observation. Furthermore, it is also noticeable that models with higher orders require fewer steps to restore the image, consistent with our discussion in \cref{method:complexity}. \cref{fig:visual_cdd} displays the visual comparison of various methods on CDD-11. It is apparent that methods integrated with CoR exhibit comparable, or even superior, visual performance compared to end-to-end methods. Notably, even the 1-order OneRestore achieves satisfactory performance with CoR. However, it is also evident that the 1-order HAIR performs poorly due to Degradation Coupling, as detailed in \cref{sec:limitation}. See \cref{sup:experiment} in supplementary  for more results.\vspace{-0.2cm}

\vspace{-0cm} \subsection{Ablation Study}
\textbf{Ablation Study of Degradation Discriminator (DD).} As shown in \cref{tab:abl_DD}, we investigate the influence of the proposed DD across five distinct settings: 
\textit{(a)} Without the DD, at each step, the model either randomly selects a degradation basis or halts the restoration process.
\textit{(b)} Only a classifier is employed each time to determine the degradation basis or to decide when to stop the restoration process. So $\mathbf{v}_i'=\mathbf{v}_i$ in \cref{eq:update_v}.
\textit{(c)} The model incorporates $\epsilon_o$. 
\textit{(d)} The model incorporates $\epsilon_{b_i}$. 
\textit{(e)} The model utilizes both $\epsilon_o$ and $\epsilon_{b_i}$. 
The significance of $\epsilon_o$ is evident in its ability to prioritize degradation bases of higher orders. Similarly, $\epsilon_{b_i}$ underscores the value of selecting an appropriate restoration sequence, as illustrated in \cref{fig:order}. Collectively, these findings validate the rationale behind our design of the DD.

\begin{table}[h]
    \centering
        \centering
        \scriptsize
        \fboxsep0.8pt
        \setlength\tabcolsep{8pt}
        \captionof{table}{{Impact of key components Degradation Discriminator.} Results are from low-light+haze+snow task on CDD-11 using a pre-trained 2-order OneRestore model. }
        \label{tab:abl_DD}
    \begin{tabularx}{\linewidth}{X*{5}{c}}
            \toprule
                Setting &  \textit{Classifier}&  $\epsilon_o$ & $\epsilon_{b_i}$ & PSNR & SSIM \\
             \midrule

            \textit{(a)} & \xmark & \xmark & \xmark  & 15.02 & 0.5137\\
            \textit{(b)} & \cmark & \xmark & \xmark  & 19.69 & 0.7109\\
            \textit{(c)} & \cmark &  \cmark & \xmark  & 21.15 & 0.7388 \\
            \textit{(d)} & \cmark & \xmark & \cmark  &  22.36 & 0.7482\\
             \textit{(e) (ours)} & \cmark & \cmark & \cmark & \textcolor{red}{24.68} & \textcolor{red}{0.7558} \\
             \bottomrule
        \end{tabularx}
   
\end{table}

\textbf{Effect of Bases.} As discussed in \cref{sec:method}, a fundamental assumption of CoR is that the composite degradations encountered must be directly combinable from the bases on which the models are trained. This concept is vlidated in \cref{tab:abl_bases}. In setting \textit{(a)}, the input images cannot be restored because the selected bases lack the "low" component. Setting \textit{(b)} includes "low," "haze," and "rain," but the combinations "rain+haze" and "low+haze" cannot be directly merged into "low+haze+rain," resulting in suboptimal outcomes. Both settings \textit{(c)} and \textit{(d)} perform well, with \textit{(d)} showing superior performance due to the inclusion of higher-order degradation bases. Selecting appropriate degradation bases is crucial for effective restoration.

\begin{table}[h]
  \centering
  \setlength{\tabcolsep}{5pt}
  \scriptsize

  \caption{Performance of the OneRestore trained under different bases with CoR on low+haze+rain of CDD-11.}
  \label{tab:abl_bases}
  \begin{tabular}{lccccccc}
    \toprule

    Setting & low & rain & haze  & rain+haze & low+haze & PSNR & SSIM\\
    \midrule
    \textit{(a)} & \xmark & \cmark & \cmark & \cmark & \xmark   &14.37 & 0.5259\\
    \textit{(b)} & \xmark & \xmark & \xmark & \cmark & \cmark   &20.03 & 0.7031 \\
    \textit{(c)} & \cmark & \cmark & \cmark & \xmark & \xmark   &22.87 & 0.7564 \\
    \textit{(d)} & \cmark & \cmark & \cmark & \cmark & \cmark   &24.57 & 0.7699 \\

    \bottomrule
  \end{tabular}\vspace{-0.5cm}
\end{table}
\par 


\section{Limitation \& Future Prospect} \label{sec:limitation}
\begin{figure}[h]
    \centering
    \includegraphics[width=0.8\linewidth]{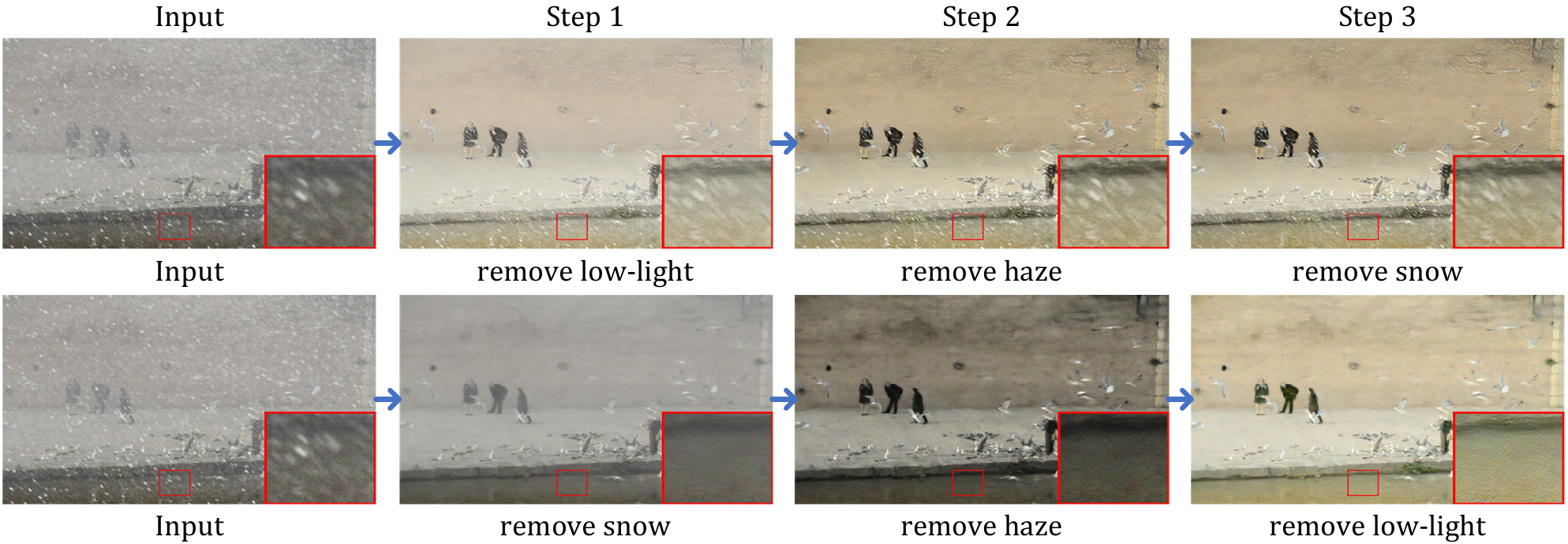}
    \caption{The visualization of the process of CoR in different restoration sequences using the same pre-trained OneRestore.}\vspace{-0.5cm}
    \label{fig:order}
\end{figure}

Before delving into the limitations, it is crucial to define a key concept: \textbf{Degradation Coupling}. Consider input images $x \sim p(x|s_1,s_2,\cdots,s_n)$, where $p(x|s_1,s_2,\cdots,s_n)$ represents the distribution of images subjected to composite degradations $s_1+s_2+\cdots+s_n$. Let $\mathbf{M}$ be a model trained on degradation bases ${b_1,b_2,\cdots,b_m}$. Suppose that $s_1+s_2+\cdots+s_n$ can be expressed as $b_{k_1}+b_{k_2}+\cdots+b_{k_t}$, and the model $\mathbf{M}$ sequentially removes these degradations starting with $b_{k_1}$. If the following condition holds:
\begin{equation}
    p(x|\mathbf{M}[b_{k_1}],b_{k_2},\cdots,b_{k_t}) \neq p(x|b_{k_2},\cdots,b_{k_t})
\end{equation}
we term this phenomenon \textbf{Degradation Coupling}. Here, $\mathbf{M}[b_{k_1}]$ indicates that the model $\mathbf{M}$ has been applied to eliminate the degradation $b_{k_1}$ from $x$. In essence, Degradation Coupling occurs when the model's removal of one degradation inadvertently affects other unintended degradations. This issue arises because the model wasn't trained to handle the scenario where \(p(x|\mathbf{M}[b_{k_1}],b_{k_2},\cdots,b_{k_t})\). For example, as depicted in \cref{fig:order}, when the model first attempts to remove low-light conditions, it inadvertently enhances other degradations like snow and haze, which it hasn't learned to address. Conversely, if the model first removes snow, which affects other degradations less, it can then more effectively restore the image. Tables \ref{tab:uird12} and \ref{tab:cdd11} show CoR performs better with non-blind methods, which can control the removal of one degradation at a time, thus reducing Degradation Coupling. Therefore, CoR is more suitable for non-blind methods. 
\par
Given the inherent interdependencies among degradations in composite degradations, Degradation Coupling is an unavoidable issue. While adjusting the restoration sequence may offer some relief, its effectiveness is limited. The direction of future works should be to develop strategies that (1) enable the model to clearly remove the targeted degradation $b_i$, and (2) minimize the impact on other degradations.  We look forward to future work that can develop algorithms to more effectively address this limitation.

\section{Conclusion}
In this paper, we introduce a new task setting called Universal Image Restoration (UIR) to address the limitations of previous settings. UIR challenges the model to be trained on a set of degradation bases and then tested on images with isolated or composite degradations in a zero-shot manner. To meet this challenge, we propose the first algorithm for UIR, known as Chain-of-Restoration (CoR). CoR enhances a pre-trained multi-task model with a Degradation Discriminator, enabling it to remove one degradation basis at a time and restore degraded images step by step. Extensive experiments indicate that CoR significantly boosts model performance in removing composite degradations, rivaling or even outperforming end-to-end methods trained on all degradations, as demonstrated by both quantitative and visual results. Lastly, we discuss the limitations and future prospects of CoR. We are confident that this work will offer new insights to the research community.


{
    \small
    \bibliographystyle{ieeenat_fullname}
    \bibliography{main.bib}
}


\clearpage
\setcounter{page}{1}
\maketitlesupplementary

\section{More Implementation \& Training Details}
\subsection{More Training Details}
\subsubsection{Multi-task Image Restoration Models.} As discussed in \cref{sec:exp}, most of the pre-trained models are directly adopted from prior works. Specifically, in \cref{tab:uird12} and \cref{tab:cdd11}, we only re-train the 1-order OneRestore model presented in \cref{tab:uird12} and the \{1,2,3\}-order HAIR, as well as the \{1,2\}-order OneRestore models detailed in \cref{tab:cdd11}. The training configurations are consistent with their original papers~\cite{hair, guo2024onerestore}.
\subsubsection{Degradation Discriminator (DD)} In the DD, we only need to train a binary/multiple image classifier. As discussed in \cref{sec:exp}, all the classifiers are \textit{mobilenet\_v3\_small} \cite{Howard_2019_ICCV} from \textit{\href{https://pytorch.org/vision/stable/index.html}{torchvision}.models}. The training dataset for the classifiers is derived from the training dataset of the image restoration models, where we simply label the training images with their respective degradation types. For blind models, we label all degraded images as "1" and clean images as "0", and then use these labels to train the binary classifier. For non-blind models, the labels on the images correspond to their degradation types, such as "0" for "haze", "1" for "rain", and so on, up to "$n+1$" for "clean", and we use these labels to train the classifier. This training process is independent of the image restoration models.

\subsection{More Implementation Details}
In our experiments, the $\epsilon_o$ is set to 0.03. For $\epsilon_{b_i}$, all degradation bases with "low-light" conditions are assigned a value of -0.05, while all others are set to 0. The primary concern is to ensure that the "low-light" degradation is addressed later in the sequence; the order of other degradations is not critical. Furthermore, due to the variable size of input images in image restoration tasks, which often differs from the size of images the classifiers were trained on, resizing the input images could introduce distortion. To mitigate this, we employ the method outlined in \cref{alg:classifier} to assist the classifier in discerning degradation types. Specifically, we randomly crop $N$ patches from the image, aggregate their information to compute the vector $\mathbf{v}$ through averaging, and subsequently derive $\mathbf{v}'$ using the update rule specified in \cref{eq:update_v}.

\begin{algorithm}[!t]
\caption{Pytorch code of how classifier discriminate the degradation type in each step.}
\label{alg:classifier}
\definecolor{codeblue}{rgb}{0.25,0.5,0.5}
\lstset{
  backgroundcolor=\color{white},
  basicstyle=\fontsize{6pt}{6pt}\ttfamily\selectfont,
  columns=fullflexible,
  breaklines=true,
  captionpos=b,
  commentstyle=\fontsize{7.2pt}{7.2pt}\color{codeblue},
  keywordstyle=\fontsize{7.2pt}{7.2pt},
}
\begin{lstlisting}[language=python]
# X: [H,W,C] input image with degradation
# cls: The image classifier to discriminate degradations
# patch_size: the size of images the classifiers are trained on
# N: the number of iterations
# v: the output probability vector
def get_patch(X,patch_size):
    # random crop a patch from 
    H,W=image.shape[0],image[1]
    h,w=random.randint(0,H-patch_size),random.randint(0,W-patch_size)
    return img[...,h:h+patch_size,w:w+patch_size]

def detect_deg(X,cls,N=12):
    v=0
    for i in range(N):
        v+=cls(get_patch(X))
    return v/N

\end{lstlisting}
\end{algorithm}

\section{More Experiments Results} \label{sup:experiment}
\begin{figure}[h]
      \centering
    \begin{subfigure}{\linewidth}
        \centering
        \includegraphics[width=1\linewidth]{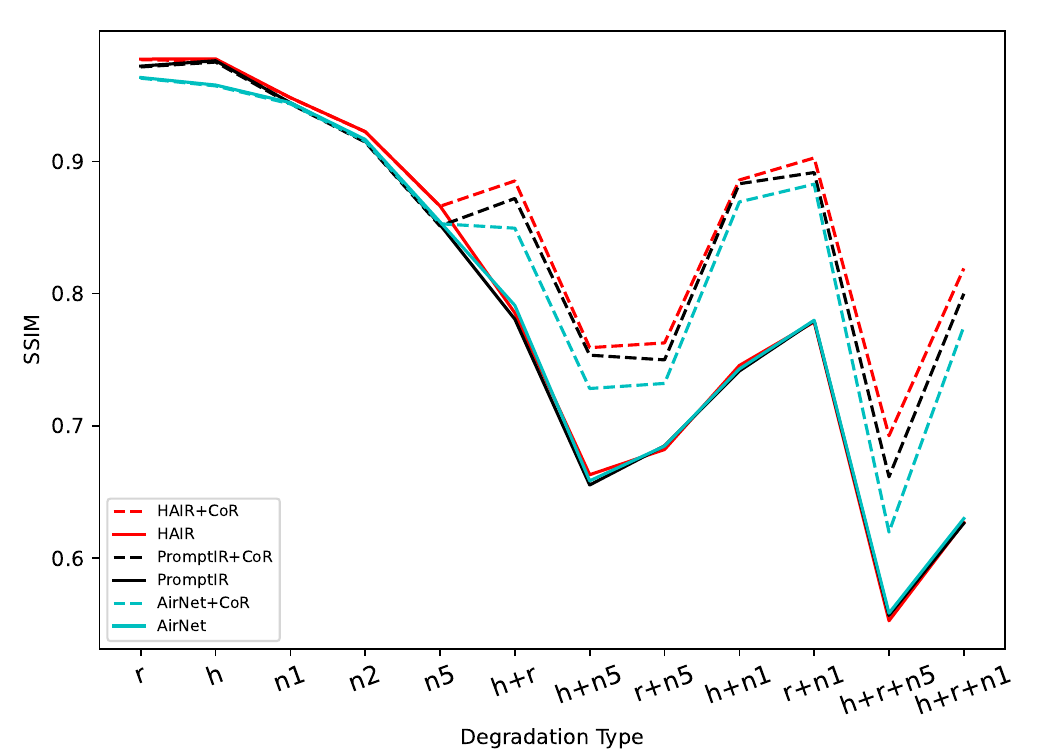}
        \caption{CoR for blind methods.}
        \label{fig:uird_blind_SSIM}
    \end{subfigure}%
    \vfill
    \begin{subfigure}{\linewidth}
        \centering
        \includegraphics[width=1\linewidth]{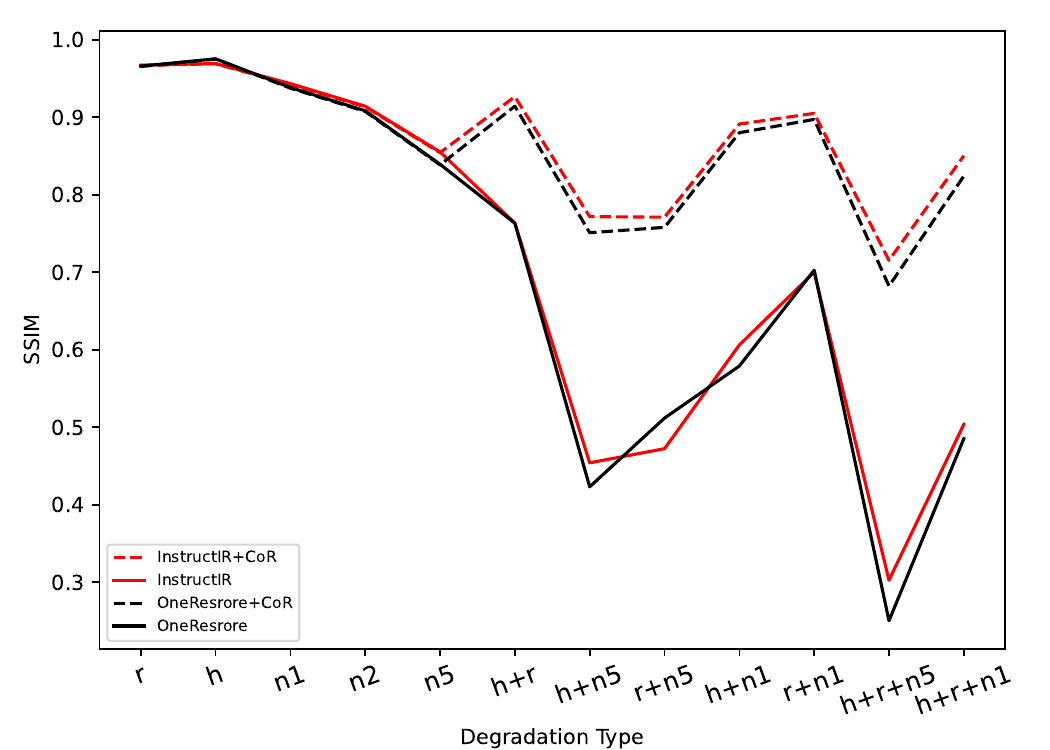}
        \caption{CoR for non-blind methods.}
        \label{fig:uird_nonblind_SSIM}
    \end{subfigure}%

    \vspace{-0cm}\caption{Visual Comparison (SSIM) of Multi-Task Models' Performance with and without CoR on UIRD-12.}\vspace{-0cm}
        \label{fig:uird12_SSIM}
\end{figure}

We provide more visualization of performance of different methods on UIRD-12 and CDD-11 in \cref{fig:uird12_SSIM} and \cref{fig:cdd11_psnr}, which also shows what we claim in our main paper.
\par
\begin{figure*}[b]
    \centering
    \includegraphics[width=1\linewidth]{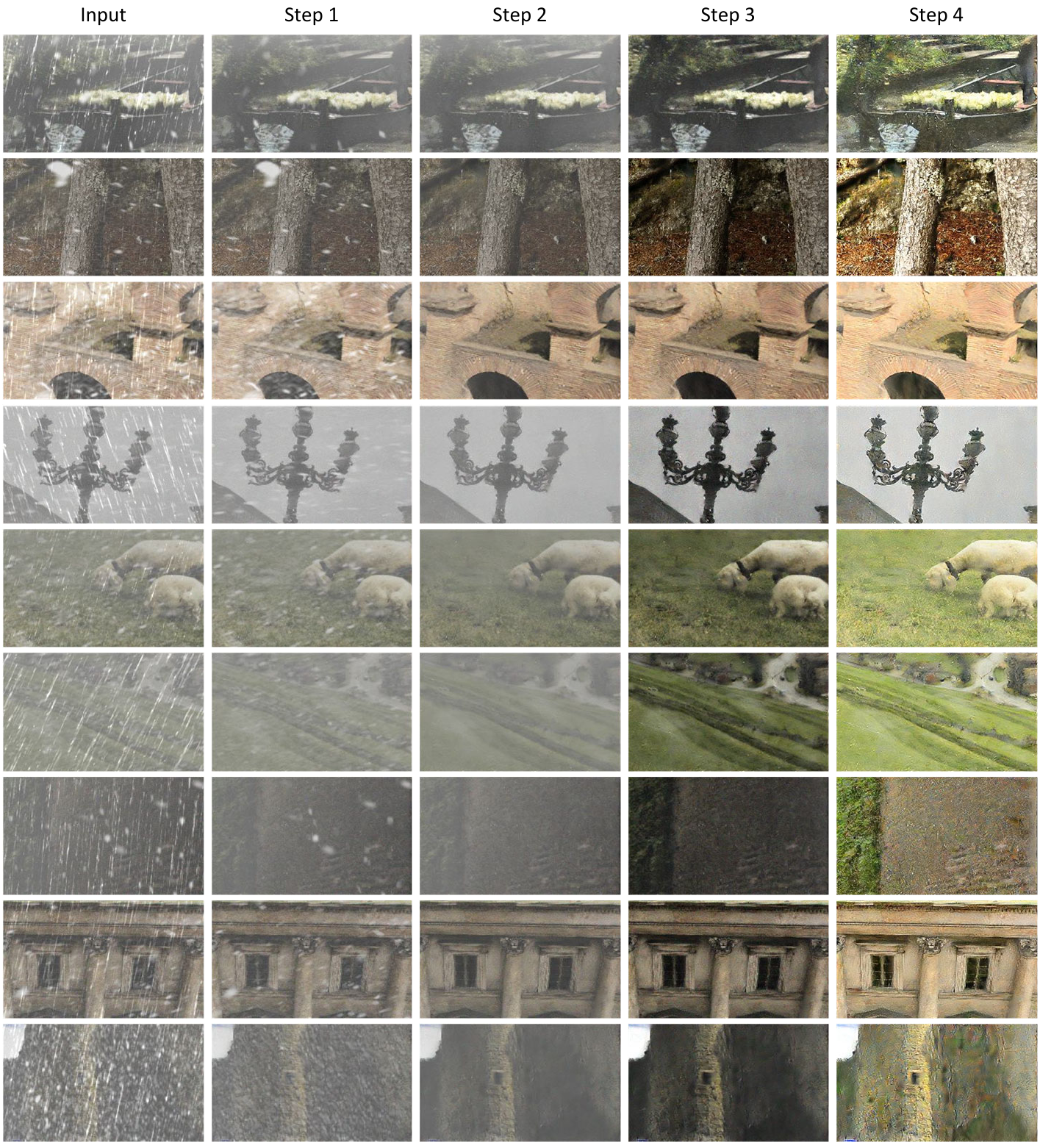}
    \caption{Visualization of step-by-step degradation removal on images with \textit{low+snow+rain+haze} using 1-order OneRestore.}
    \label{fig:visual_low_snow_rain_haze}
\end{figure*}
\begin{figure*}[b]
    \centering
    \includegraphics[width=1\linewidth]{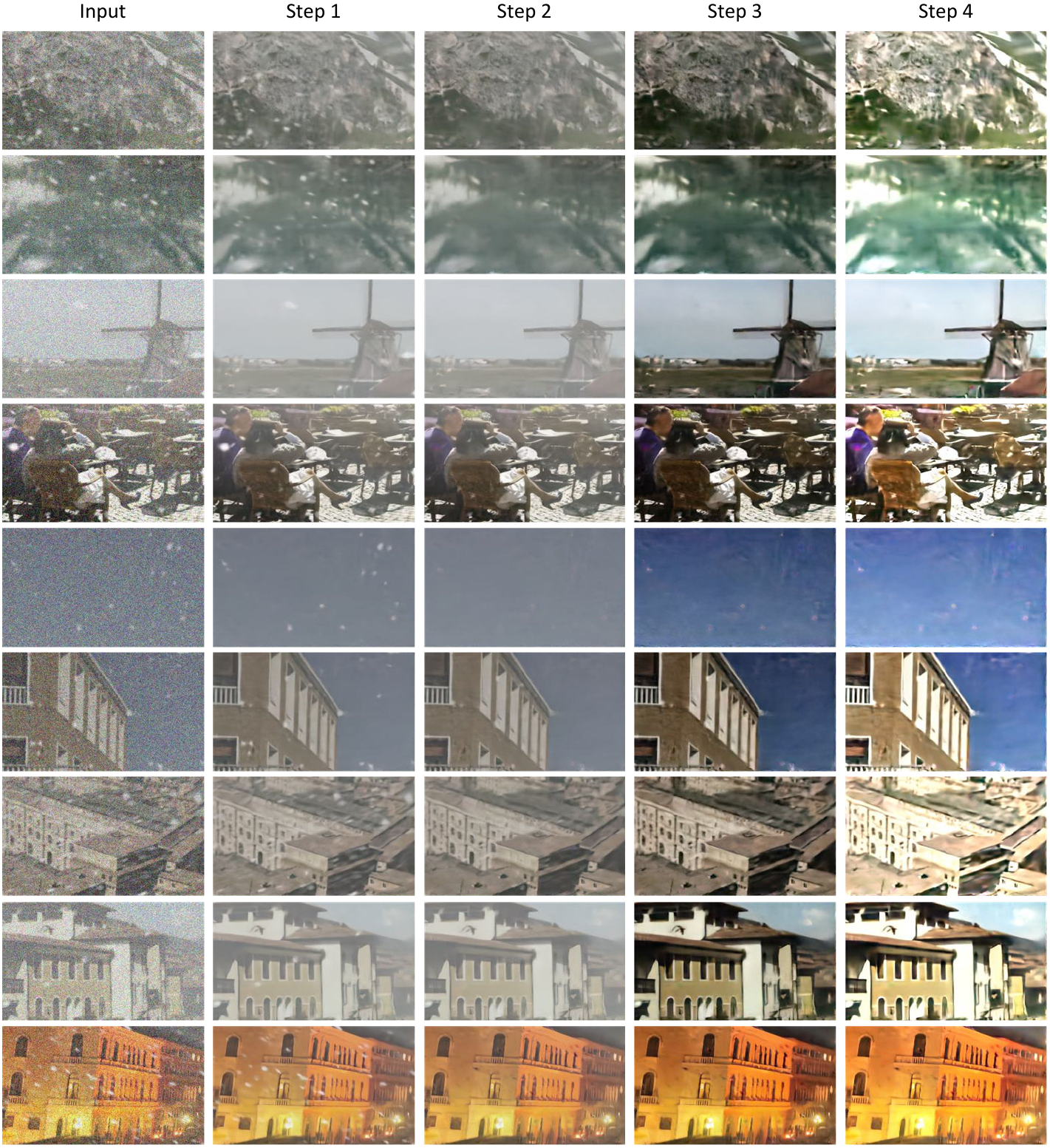}
    \caption{Visualization of step-by-step degradation removal on images with \textit{low+snow+noise($\sigma$=25)+haze} using 1-order OneRestore.}
    \label{fig:visual_low_snow_noise_haze}
\end{figure*}
\begin{figure*}[b]
    \centering
    \includegraphics[width=1\linewidth]{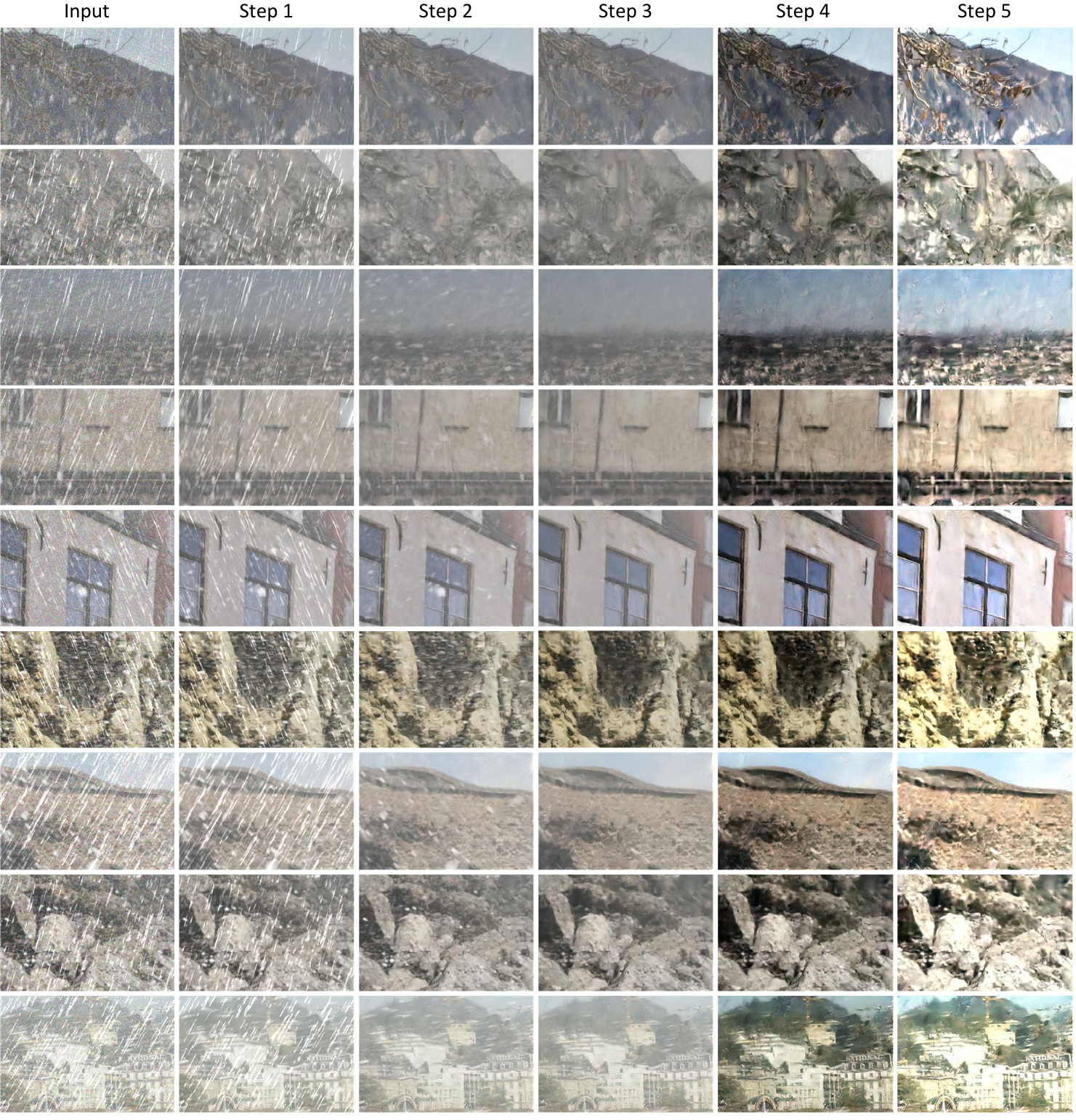}
    \caption{Visualization of step-by-step degradation removal on images with \textit{low+snow+noise($\sigma$=15)+rain+haze} using 1-order OneRestore.}
    \label{fig:visual_low_rain_snow_noise_haze}
\end{figure*}

\textbf{CoR for Degradations with Higher Orders.} In our main paper, the test dataset only contains degradations with orders $\leq 3$. Therefore, we attempt to apply CoR to degradations with higher orders, as illustrated in \cref{fig:visual_low_snow_rain_haze}, \cref{fig:visual_low_snow_noise_haze}, and \cref{fig:visual_low_rain_snow_noise_haze}. It is evident that CoR performs robustly across various complex composite degradations. However, it is also clear that as the degradations increase in complexity and intensity, the issue of degradation coupling becomes more pronounced. For example, in \cref{fig:visual_low_snow_noise_haze}, we can observe that after the removal of Gaussian noise, other degradations become "fuzzy," which prevents the model from completely eliminating the remaining degradations. This scenario highlights that, despite being the best restoration sequence, the effect of altering the restoration order is limited. Therefore, we should focus more on designing a model that can eliminate a degradation with minimal impact on other degradations.

\section{Discussion}

\subsection{The Impact of Classifier is Relatively Small}
In addition to Degradation Coupling, another factor that can decrease the performance of CoR is the incorrect discrimination made by the Classifier. For instance, in \cref{tab:uird12}, it can be observed that the performance of multi-task models on isolated degradations decreases, which is attributed to some errors made by the Classifier. However, since discriminating dozens of degradation types is considerably simpler compared to image classifications involving hundreds of classes, the error rate is generally acceptable ($\leq 5\%$). In our main paper, we chose MobileNet~\cite{Howard_2019_ICCV} as the Classifier in the Degradation Discriminator (DD), and we found that when we changed it to other backbones such as~\cite{Simonyan2014VeryDC,resnet,Liu_2021_ICCV,Huang2016DenselyCC}, the performance of CoR remained nearly the same. This indicates that our focus should be more on the multi-task model itself rather than the choice of the Classifier.
\subsection{The Sequence of Restoration Without Control.}
As described in \cref{sec:method:DD}, for blind models, we employ a binary classifier solely to determine if the image is clean. In contrast, for non-blind models, we utilize $\epsilon_o$ and $\epsilon_{b_i}$ to manage the restoration sequence. If we eliminate $\epsilon_o$ and $\epsilon_{b_i}$, the approach becomes similar to that of blind models. Generally, without any sequence control, models with CoR tend to address the most intense degradations first. For instance, in UIRD-12, we observe the removal priority is \textit{noise} $>$ \textit{rain} $>$ \textit{haze}. Since noise is the most dense and obvious degradation, models typically target it initially. For CDD-11, the observed priority is \textit{low-light} $>$ \textit{haze} $>$ \textit{snow/rain}. This preference is due to the fact that snow and rain in CDD-11 are generally mild and sparse, whereas low-light and haze are more intense. Thus, without control, models are inclined to first remove low-light, which can lead to suboptimal results, as is the case of the 1-order HAIR in \cref{tab:cdd11}.

\begin{figure}
    \centering
    \includegraphics[width=\linewidth]{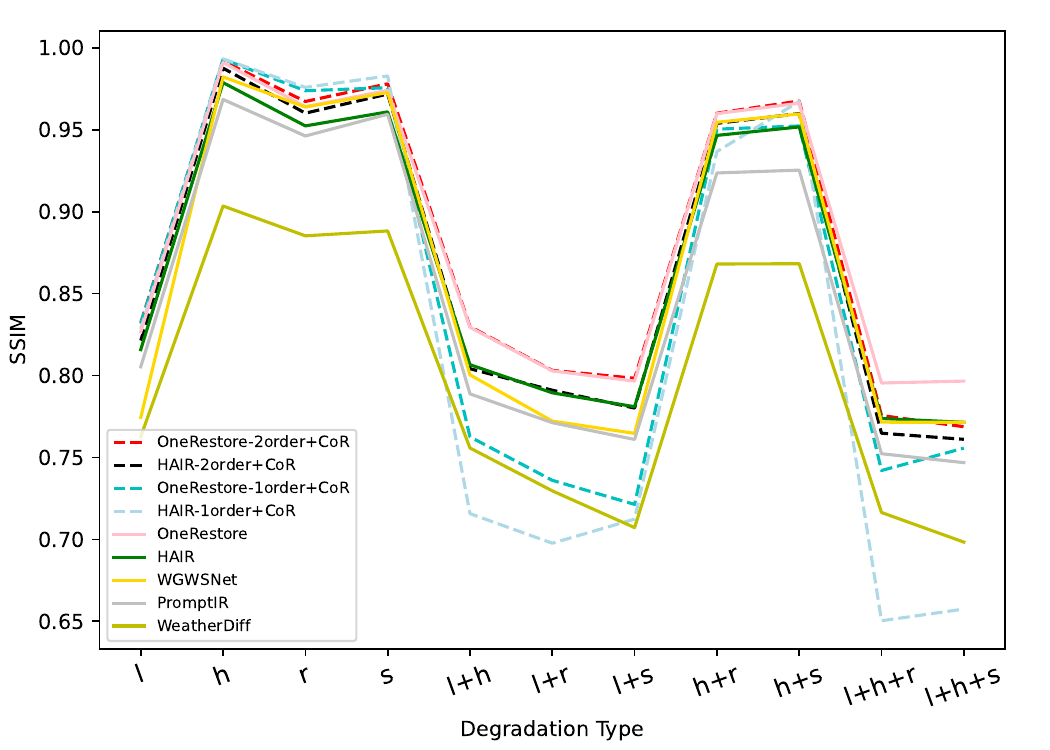}
    \caption{Visualization of performance (SSIM) of multi-task models on CDD-11. Methods without "order" are all 3-order end-to-end models trained on all degradations.}
    \label{fig:cdd11_ssim}
\end{figure}

\end{document}